# Title

Memorization in Large Language Models in Medicine Prevalence Characteristics and Implications

**Author list**

Anran Li[1], Lingfei Qian[1], Mengmeng Du[2], Yu Yin[3], Yan Hu[4], Zihao Sun[5], Yihang Fu[2], Hyunjae Kim[1], Erica Stutz[1], Xuguang Ai[1], Qianqian Xie[1], Rui Zhu[1], Jimin Huang[1], Yifan Yang[6], Siru Liu[7,8], Yih-Chung Tham[9], Lucila Ohno-Machado[1], Hyunghoon Cho[1], Zhiyong Lu[6], Hua Xu[1], Qingyu Chen[1,*]

**Corresponding author**:

[*]Qingyu Chen, qingyu.chen@yale.edu

**Affiliations**

1. Department of Biomedical Informatics and Data Science, School of Medicine, Yale University, New Haven, USA
2. Health Informatics, Yale School of Public Health, Yale University, New Haven, USA
3. Department of Earth Science and Engineering, Imperial College London, London, United Kingdom
4. McWilliams School of Biomedical Informatics, University of Texas Health Science at Houston, Houston, USA
5. University of California, San Diego, USA
6. National Library of Medicine, National Institutes of Health, Maryland, USA
7. Department of Biomedical Informatics, Vanderbilt University Medical Center, Nashville, USA
8. Department of Computer Science, Vanderbilt University, Nashville, USA
9. Department of Ophthalmology, Yong Loo Lin School of Medicine, National University of Singapore, Singapore

# 1. Abstract

Large Language Models (LLMs) have demonstrated significant potential in medicine, with many studies adapting them through continued pretraining or fine-tuning on medical data. However, a key question remains: to what extent do LLMs memorize medical training data—that is, recall or regenerate content seen during continued pretraining or fine-tuning.

In this work, we investigate memorization of LLMs in medicine, assessing its prevalence (frequency), characteristics (what is memorized), volume (how much), and potential downstream impacts. We systematically analyze common adaptation scenarios: (1) continued pretraining on medical corpora, (2) fine-tuning on standard medical benchmarks, and (3) fine-tuning on real-world clinical data, including over 13,000 unique inpatient records from Yale New Haven Health System. The results demonstrate that memorization is prevalent and significantly higher than that in the general domain. Memorization has distinct characteristics during continued pretraining and fine-tuning, and it is persistent: up to 87% of content memorized during continued pretraining remains after fine-tuning. Memorization can be categorized into three types: beneficial (e.g., accurate recall of clinical guidelines), uninformative (e.g., templated language), and harmful (e.g., sensitive clinical content). We offer practical recommendations to facilitate beneficial memorization, minimize uninformative memorization, and mitigate harmful memorization to protect patient privacy and improve medical utility.

## 2. Introduction

Large language models (LLMs) represent a significant advancement in foundation models [1,2]. LLMs demonstrate superior performance particularly in generative applications such as question answering [3,4] and text summarization [5] and show more robust capabilities in zero- and few-shot settings [6,7]. Despite these advances, studies have reported mixed results on LLM performance in specialized domains such as medicine [8–11]. Some research highlights promising performance in specific clinical tasks, such as extracting findings from radiology reports [12], while other studies have shown that LLMs may underperform in tasks like clinical information extraction [12], medical document classification [13], and disease diagnosis [13], and may introduce risks such as diagnostic errors [13] and factual hallucinations [14]. Systematic evaluations of LLMs in medicine further suggest that direct applying these models could result in inconsistencies, missing information, and hallucinated content [7,15]. A key limitation is that general-purpose LLMs often lack domain-specific knowledge and reasoning capabilities, which may undermine their accuracy and safety in medical applications [16].

To address these challenges, studies have adopted continued pretraining or supervised fine-tuning on medical data to adapt LLMs for use in medicine [8,13,17–21]. For instance, multiple foundation language models (e.g., PMC-LLaMA [17], Meditron [18], Me-LLaMA [19], Med-Llama-3 [22], and gpt-oss [23]) have been developed through continued pretraining, where general-purpose LLMs (e.g., LLaMA) are further pretrained on large-scale medical corpora (e.g., biomedical literature and clinical notes) using self-supervised learning to capture domain-specific knowledge. In parallel, many studies fine-tune LLMs on labeled medical datasets through instruction tuning to improve their ability to follow domain-specific instructions. Applications include fine-tuning for medical record summarization [24], diagnostic reasoning [13], clinical record correction [8], and early disease group predictions [20], using both medical foundation models and general-purpose LLMs. These approaches have shown improved effectiveness in medical applications while reducing hallucinations and enhancing safety [11]. For example, studies have shown that adapted models can outperform directly applying LLMs—including state-of-the-art closed-source LLMs—by up to 30% on clinical information extraction [25] and 20% on disease diagnosis [26].

Despite their advances, a key question remains: to what extent do LLMs memorize medical training data during domain-specific adaptation? Memorization refers to when models recall or regenerate content seen during continued pretraining or fine-tuning. On one hand, memorization may be beneficial if it allows LLMs to retain valuable medical knowledge—such as terminology, clinical guidelines, or biomedical references—as intended by the adaptation process. On the other hand, memorization in the medical domain introduces critical risks. First,

LLMs may inadvertently reproduce sensitive patient-specific information, posing serious privacy concerns [27]. Second, the ultimate goal of domain adaptation is to help LLMs capture medical knowledge and reasoning so they can adapt to diverse medical applications—not simply repeat training content. Excessive memorization may indicate that a model is relying on surface-level copying rather than acquiring deeper medical understanding, which can limit its generalizability [28,29]. These concerns are further amplified by the generative nature of LLMs [3,30,31], which may lead to unintended disclosures or unwarranted outputs, ultimately hindering their adoption in medicine.

In this study, we systematically examine the memorization of LLMs in medicine, focusing on three core aspects: (1) prevalence—how often content is memorized; (2) characteristics—what types of content are memorized; (3) volume—how much content is memorized; and (4) downstream impact—how memorization affects medical applications. We evaluate memorization across three common adaption settings: (1) continued pretraining over medical corpora, (2) fine-tuning over medical benchmarks, and (3) fine-tuning over real-world clinical data. Our analysis spans both medical foundation models (PMC-LLaMA, Meditron, Me-LLaMA, and Med-Llama-3 variants) and general-purpose LLMs (Llama 2 and 3 variants and gpt-oss), covering ten datasets comprising hundreds of thousands of instances and extensive manual review of thousands of model outputs. For fine-tuning on real-world clinical data, we conduct a case study on LLM-assisted disease diagnosis using 13,000 unique inpatient records from the Yale New Haven Health System, partitioned into 10,000 for training, 1,000 for validation and 2,000 for testing.

First, memorization is prevalent across all adaptation scenarios. For instance, during continued pretraining, memorization ratios of consecutive 30 tokens range from 10% to 20% across input lengths on selected medical corpora. In fine-tuning on medical question answering (QA) benchmarks, models regenerate 14% to 21% of removed answer options. In fine-tuning on clinical data, while fine-tuned models improve diagnostic accuracy, they also have 2%–30% memorization ratios of consecutive 30 tokens and regenerate 3,192 Protected Health Information (PHI) [32,33] instances among the 10,000 training records. Manual review of 200 outputs further reveal 98 cases of potentially sensitive information, e.g., highly sensitive diagnoses, family relationships, and treatment details, beyond standard PHI definitions [32–34].

In addition, LLMs also have distinct memorization types that can be broadly categorized into three groups based on their impact. First, beneficial memorization involves the accurate recall of domain-specific knowledge, such as biomedical concepts and clinical guidelines, after domain adaptation; this type is desirable as it supports domain-specific reasoning and accuracy. Second, uninformative memorization refers to the regeneration of templated language, including standard disclaimers and repetitive statements, which reflects surface-level copying rather than genuine acquisition of domain-specific knowledge or reasoning skills. Third, harmful memorization occurs when models reproduce data-specific content and potentially sensitive clinical information, such as reproducing removed answer options when fine-tuned on medical QA benchmarks or PHI when fine-tuned on clinical data. A unique characteristic of memorization in LLMs for medicine is its persistence: up to 87% of content memorized during continued pretraining remains after fine-tuning on new medical tasks. Memorization characteristics also vary by adaptation stage; for example, models fine-tuned on medical benchmarks tend to have lower rates of consecutive token matches but still display significant dataset-specific memorization. Furthermore, results show that memorization increases with model size and input length, while decoding parameters such as temperature have limited influence.

We further provide practical recommendations addressing both the positive and negative implications of memorization for the development and adoption of LLMs in medicine. We also call for community-wide efforts to establish improved reporting guidelines and to incorporate memorization awareness into future deployment frameworks. To support reproducibility and future advancement, we publicly release the full codebase and models trained on public datasets.

# 3. Results

## 3.1 Memorization results of continued pretrained LLMs

Table 2 presents memorization results across the six evaluation metrics during the continued pretraining stage with an input length $l$=50. Figure 1 further illustrates the results for consecutive 30-token and 50-token ratio across varying input lengths ($l$ = 50, 100, 200, 300, and 500). Detailed results including additional measures and statistical analysis are provided in the Supplementary Material Section S2. Note that results for PMC-LLaMA are excluded, as the model frequently failed to follow the prompts and instead produced hallucinated and repetitive responses in this specific setting. As a result, its outputs were not representative for evaluation (details are provided in Supplementary Material Section S2, with examples provided in Section S5).

**Comparisons with baselines**. Overall, the results show that medical foundation language models have higher memorization ratios than their corresponding baselines. For example, on the Clinical guidelines dataset, Meditron 7B achieves a 30-token memorization ratio of 10.48%, compared to 1.23% for its direct baseline Llama 2 ($p < 0.0001$) and 1.30% for LLaMA ($p < 0.0001$). A similar trend is observed for Med-Llama-3 compared to Llama 3 on the same dataset. On the MIMIC-III dataset, Me-LLaMA reaches a 30-token memorization ratio of 15.32%, whereas Llama 2 achieved only 0.10% ($p < 0.0001$). In contrast, the difference in memorization on biomedical literature datasets is smaller – though still consistently higher for the medical models. For instance, on the PubMed Central (PMC) full-text dataset, Meditron 7B achieves 0.25%, compared to 0.18% for Llama 2 ($p <0.0002$) and 0.11% for LLaMA ($p <0.0002$). An exception is noted for the MEDLINE abstracts dataset, where Meditron 7B and Llama 2 show similar memorization ratios: 0.50% vs. 0.51% for 30-token spans, and 0.33% vs. 0.29% for 50-token spans. This might indicate that Llama 2's pretraining corpus includes MEDLINE abstracts. Studies report minimal performance gains when comparing Meditron and Llama 2 on biomedical datasets curated from MEDLINE abstracts, including findings from the Meditron study itself [18].

Table 2. Detailed memorization results during continued pretraining with input length $l$ = 50. The results are reported using bootstrapping with a random sample of 100 instances, repeated 100 times for each dataset. LLaMA is provided as an additional reference.

| Datasets | Models | Exact measures | | Approximate measures | | Semantic measures | |
|---|---|---|---|---|---|---|---|
| | | Consecutive 30 tokens | Consecutive 50 tokens | ROUGE-L | BLEU | BERT | BART |
| **Meditron** | | | | | | | |
| Clinical guidelines | Llama-2-7B (Baseline) | 1.23% | 0.55% | 0.200 | 0.047 | 0.668 | -4.776 |
| | LLaMA-7B (Baseline) | 1.30% | 0.63% | 0.201 | 0.046 | 0.671 | -4.828 |
| | Meditron-7B | 10.48% | 8.85% | 0.279 | 0.126 | 0.711 | -4.369 |
| | Meditron-70B | **21.78%** | **14.35%** | **0.360** | **0.220** | **0.750** | **-3.985** |
| MEDLINE abstracts | Llama-2-7B (Baseline) | 0.51% | 0.29% | 0.184 | 0.023 | 0.664 | -4.851 |
| | LLaMA-7B (Baseline) | 0.35% | 0.23% | 0.179 | 0.024 | 0.672 | -4.882 |
| | Meditron-7B | 0.50% | 0.33% | 0.194 | 0.033 | 0.680 | -4.759 |
| | Meditron-70B | **0.59%** | **0.33%** | **0.213** | **0.043** | **0.701** | **-4.591** |
| PMC full-text articles | Llama-2-7B (Baseline) | 0.18% | 0.05% | 0.144 | 0.017 | 0.651 | -5.034 |
| | LLaMA-7B (Baseline) | 0.11% | 0.06% | 0.157 | 0.020 | 0.660 | -4.951 |
| | Meditron-7B | 0.25% | 0.11% | 0.178 | 0.029 | 0.670 | -4.787 |
| | Meditron-70B | **0.41%** | **0.17%** | **0.189** | **0.037** | **0.682** | **-4.695** |
| Experience reply | Llama-2-7B (Baseline) | 2.27% | 1.17% | 0.176 | 0.055 | 0.635 | -5.456 |
| | LLaMA-7B (Baseline) | 1.80% | 0.88% | 0.168 | 0.049 | 0.634 | -5.501 |
| | Meditron-7B | 1.95% | 0.92% | 0.169 | 0.052 | 0.636 | -5.475 |
| | Meditron-70B | **3.68%** | **1.84%** | **0.200** | **0.075** | **0.661** | **-5.257** |
| **Med-Llama-3** | | | | | | | |
| Clinical guidelines | Llama-3-8B (Baseline) | 5.77% | 2.22% | 0.257 | 0.099 | 0.711 | -4.528 |
| | LLaMA-7B (Baseline) | 1.30% | 0.63% | 0.201 | 0.046 | 0.671 | -4.828 |
| | Med-Llama-3-8B | **11.33%** | **8.80%** | **0.296** | **0.140** | **0.728** | **-4.339** |
| MEDLINE abstracts | Llama-3-8B (Baseline) | 0.43% | 0.24% | 0.198 | 0.032 | 0.694 | -4.747 |
| | LLaMA-7B (Baseline) | 0.35% | 0.23% | 0.179 | 0.024 | 0.672 | -4.882 |
| | Med-Llama-3-8B | **0.47%** | **0.28%** | **0.209** | **0.037** | **0.697** | **-4.719** |
| **Me-LLaMA** | | | | | | | |
| MIMIC-III | Llama-2-7B (Baseline) | 0.10% | 0.02% | 0.056 | 0.012 | 0.582 | -6.855 |
| | LLaMA-7B (Baseline) | 0.24% | 0.08% | 0.058 | 0.012 | 0.594 | -6.690 |
| | Me-LLaMA-13B | **15.32%** | **5.41%** | **0.335** | **0.327** | **0.783** | **-4.945** |

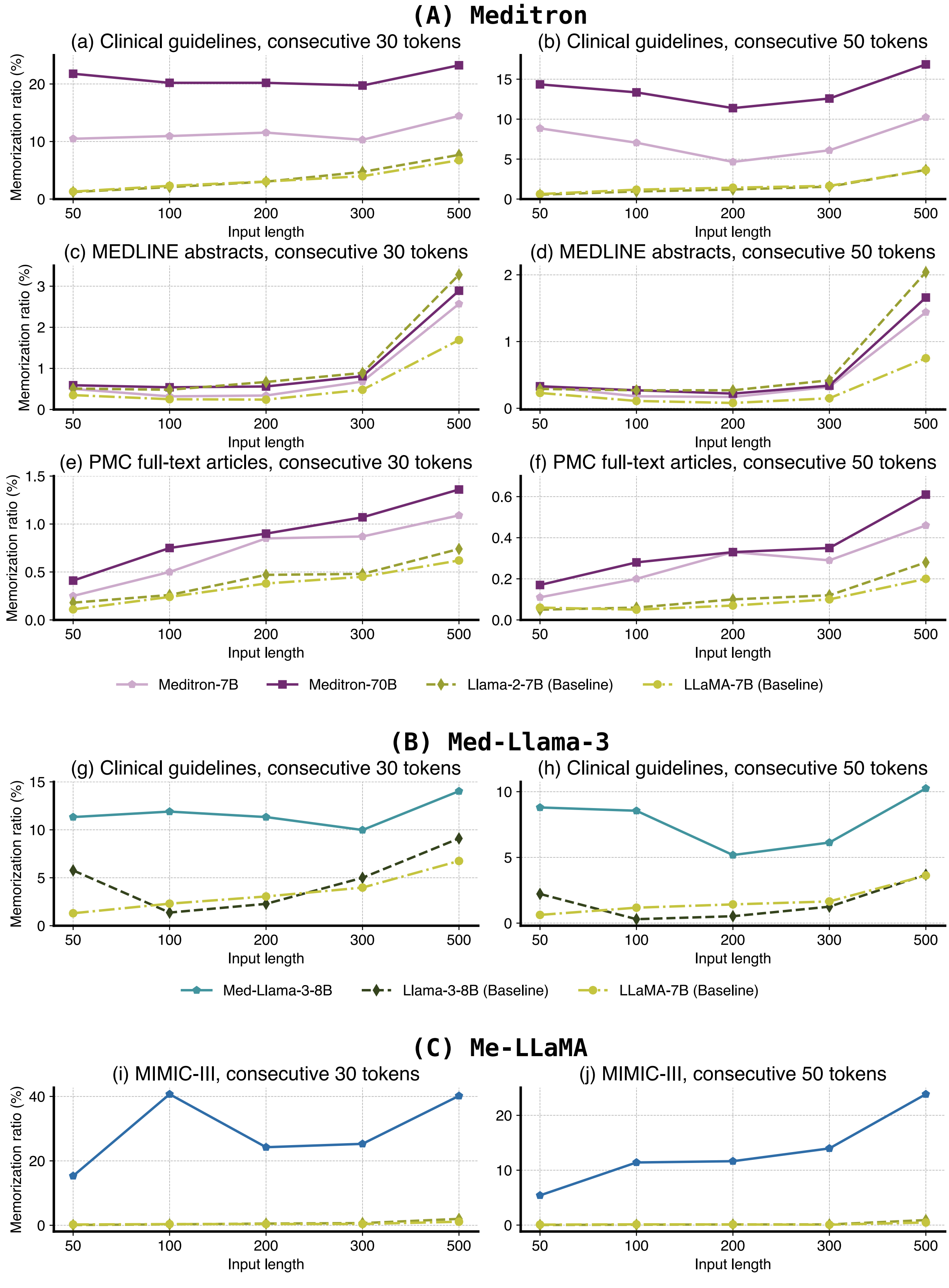

Figure 1. Results of 30-token and 50 token-memorization ratios across different input lengths.

**Position analysis on where memorization occurs.** We further analyze where memorization occurs in LLM responses, using Meditron-7B on the Clinical guidelines dataset as an example. Specifically, we examine the positions of consecutive 30-token matches. As the model is prompted to generate up to 500 tokens, we divide the generated text into 100-token segments (e.g., first 100 tokens, 101-200 tokens, etc.). Figure S2.2 (Supplementary Material Section S2.1) illustrates the distribution of memorized text across these different segments. With varying input lengths $l$=50, 100, the results consistently show that memorized text predominantly appeared within the first 100 tokens, with its prevalence steadily decreasing in subsequent segments. For instance, on the Clinical guidelines dataset corpus at $l$=50, Meditron-7B has an 30-token memorization ratio of 10.48%, with 8.38% of the memorized text occurring within the first 100 tokens, 0.75% within the 101-200 token range, and significantly less in the remaining segments.

**Manual examination on what is memorized**. We further conduct a manual examination on 800 consecutive 50 token samples from the Meditron-7B responses. As it has more instances in Clinical guidelines, we randomly sample 400 consecutive 50 token instances from Clinical guidelines and 200 instances each for MEDLINE abstracts and PMC full-text articles. These samples are manually reviewed and categorized based on its content, with the results summarized in Table S2.2 (Supplementary Material Section S2.1). The analysis shows that the model recalled useful and relevant medical information, including drug-related content (118 instances), such as medication instructions and adverse events; patient care guidance (39 instances), such as patient presentations and management strategies; and biomedical concepts (28 instances), such as gene and protein information. These results suggest that models can retain meaningful medical knowledge following domain adaptation. In contrast, the model also reproduces boilerplate or templated language. For instance, we find 192 instances of identical disclaimers in the Clinical guidelines corpus, such as: "WikiDoc is intended to be an educational tool, not a tool for any form of healthcare delivery. The educational content on WikiDoc drug pages is based upon the FDA package insert, National Library of Medicine content, and practice guidelines/consensus statements." Such surface-level memorization contributes little to domain- or task-specific knowledge.

**Additional approximate and semantic measures.** As introduced earlier, we also employ approximate and semantic measures as additional measures to complement stringent exact measures. As shown in Table 2, these additional metrics also demonstrate that medical foundation language models have higher memorization ratios than their corresponding baselines. We further compare memorization ratios across approximate and semantic measures in Supplementary Material Section S2.1, which reveals up to a two-fold increase in potentially memorized instances relative to exact measures. Table S2.3 (Supplementary Material Section S2.1) presents some examples identified by approximate measures. Supplementary Material Section S2.3 presents more representative examples—cases that do not meet the strict consecutive 30- or 50-token thresholds but include nearly identical substrings (sometimes spanning multiple consecutive sentences) or convey the same meaning using different word order or phrasing.

Table 3. Memorization results of the large language models (LLMs) fine-tuned over standard benchmarks, the training sets of MedQA and MedMCQA. LLaMA is provided as an additional reference baseline because its fine-tuned datasets are publicly accessible and do not overlap with the benchmarks whereas the fine-tuned datasets of other baseline models are not publicly available. For primary LLMs, we use the fine-tuned models on MedQA and MedMCQA made available by the original studies where possible. For Meditron and Med-Llama-3, as the fine-tuned versions are not available by the original studies, we fine-tune and make the models and training details available. PMC-LLaMA-13B (Original): the fine-tuned PMC-LLaMA-13B provided by the original study. PMC-LLaMA-13B (Reproduced): we fine-tune PMC-LLaMA-13B locally for verification. Exact, approximate, and semantic memorization measures are reported with input lengths $l$ = 50. Further results including additional input lengths and statistical analysis are provided in Supplementary Material Section S3.

| **Fine-tuning datasets** | **Model** | **Exact measures** | | **Answer option regeneration measure** | | **Approximate measures** | | **Semantic measures** | |
|---|---|---|---|---|---|---|---|---|---|
| | | **Consecutive 30 tokens** | **Consecutive 50 tokens** | **Exact Match** | **Relaxed Match** | **ROUGE-L** | **BLEU** | **BERT** | **BART** |
| MedQA | **Primary LLMs using Llama 2 backbone comparison** | | | | | | | | |
| | Llama-2-Chat-7B (Baseline) | 3.80% | 0.10% | 5.30% | 6.20% | 0.238 | 0.085 | 0.690 | -4.166 |
| | LLaMA-Chat-7B (Baseline) | 0.00% | 0.00% | 2.80% | 3.30% | 0.186 | 0.042 | 0.697 | -4.624 |
| | **PMC-LLaMA** | | | | | | | | |
| | PMC-LLaMA-13B (Original) | **18.20%** | **4.35%** | **17.25%** | **19.00%** | **0.366** | **0.239** | **0.775** | **-3.570** |
| | PMC-LLaMA-13B (Reproduced) | 3.00% | 0.20% | 7.75% | 8.35% | 0.247 | 0.108 | 0.703 | -4.110 |
| | PMC-LLaMA-7B (Reproduced) | 2.65% | 0.15% | 6.50% | 7.15% | 0.248 | 0.110 | 0.710 | -4.070 |
| | **Meditron** | | | | | | | | |
| | Meditron-7B | 3.80% | 0.05% | 7.30% | 8.40% | 0.286 | 0.143 | 0.745 | -3.910 |
| | Meditron-70B | 4.35% | 0.15% | 9.90% | 10.50% | 0.282 | 0.142 | 0.744 | -3.905 |
| | **Me-LLaMA** | | | | | | | | |
| | Me-LLaMA-13B | 3.15% | 0.05% | 7.18% | 9.05% | 0.254 | 0.115 | 0.734 | -4.033 |
| | **Primary LLMs using Llama 3 backbone comparison** | | | | | | | | |
| | Llama-3-Instruct-8B (Baseline) | 3.10% | 0.15% | 6.45% | 8.35% | 0.154 | 0.044 | 0.697 | -3.829 |
| | LLaMA-Chat-7B (Baseline) | 0.00% | 0.00% | 2.80% | 3.30% | 0.186 | 0.042 | 0.697 | -4.624 |
| | **Med-Llama-3** | | | | | | | | |
| | Med-Llama-3-8B | **4.20%** | **0.30%** | **15.02%** | **23.72%** | **0.231** | **0.114** | **0.715** | **-3.694** |
| | **Llama-3-Fine-tuned** | | | | | | | | |
| | Llama-3-8B-Fine-tuned | 4.05% | 0.10% | 9.35% | 12.80% | 0.198 | 0.103 | 0.704 | -3.756 |
| | **Primary LLMs using gpt-oss backbone comparison** | | | | | | | | |
| | gpt-oss-20B (Baseline) | 1.70% | 0.10% | 14.15% | 16.45% | 0.139 | 0.035 | 0.682 | -3.902 |
| | gpt-oss-20B-Fine-tuned | **3.05%** | **0.15%** | **18.76%** | **20.65%** | **0.151** | **0.041** | **0.686** | **-3.895** |
| MedMCQA | **Primary LLMs using LLaMA2 backbone comparison** | | | | | | | | |
| | LLaMA2-Chat-7B (Baseline) | 0.03% | 0.00% | 12.44% | 13.58% | 0.157 | 0.025 | 0.624 | -5.510 |
| | LLaMA1-Chat-7B (Baseline) | 0.00% | 0.00% | 6.04% | 7.13% | 0.161 | 0.010 | 0.650 | -5.829 |
| | **PMC-LLaMA** | | | | | | | | |
| | PMC-LLaMA-13B (Original) | **18.92%** | **10.75%** | **18.52%** | **20.72%** | **0.419** | **0.511** | **0.858** | **-2.495** |
| | PMC-LLaMA-13B (Reproduced) | 0.01% | 0.00% | 14.22% | 15.26% | 0.184 | 0.062 | 0.658 | -5.411 |
| | PMC-LLaMA-7B (Reproduced) | 0.00% | 0.00% | 11.75% | 12.93% | 0.169 | 0.046 | 0.649 | -5.494 |
| | **Meditron** | | | | | | | | |
| | Meditron-7B | 0.12% | 0.00% | 13.70% | 14.90% | 0.195 | 0.048 | 0.678 | -5.359 |
| | Meditron-70B | 0.25% | 0.07% | 17.50% | 18.90% | 0.212 | 0.060 | 0.695 | -5.234 |
| | **Me-LLaMA** | | | | | | | | |
| | Me-LLaMA-13B | 0.03% | 0.00% | 13.59% | 15.09% | 0.187 | 0.038 | 0.666 | -5.420 |
| | **Primary LLMs using LLaMA3 backbone comparison** | | | | | | | | |
| | LLaMA3-Instruct-8B (Baseline) | 0.07% | 0.01% | 14.22% | 15.97% | 0.051 | 0.005 | 0.593 | -5.577 |
| | LLaMA1-Chat-7B (Baseline) | 0.00% | 0.00% | 6.04% | 7.13% | 0.161 | 0.010 | 0.650 | -5.829 |
| | **Med-Llama-3** | | | | | | | | |
| | Med-Llama-3-8B | 0.19% | 0.02% | **18.10%** | **30.14%** | **0.178** | **0.041** | **0.708** | **-5.245** |
| | **Llama-3-Fine-tuned** | | | | | | | | |

| | | | | | | | | |
|---|---|---|---|---|---|---|---|---|
| Llama-3-8B-Fine-tuned | **0.21%** | **0.06%** | 20.99% | 22.92% | 0.176 | 0.036 | 0.696 | -5.304 |
| **Primary LLMs using gpt-oss backbone comparison** | | | | | | | | |
| gpt-oss-20B (Baseline) | 0.07% | 0.01% | 11.27% | 16.31% | 0.054 | 0.001 | 0.500 | -6.108 |
| gpt-oss-20B-Fine-tuned | **0.11%** | **0.02%** | **16.50%** | **18.65%** | **0.080** | **0.008** | **0.612** | **-5.432** |

## 3.2 Memorization results of LLMs fine-tuned over standard medical benchmarks

**Comparison of accuracy before and after fine-tuning.** Table S2.4 (Supplementary Material Section S2.1) presents the accuracy of selected models (Meditron, Llama-3, and gpt-oss) before and after fine-tuning. Recall that only the training sets were used for fine-tuning. The accuracy was measured on the held-out datasets. Across all three models, accuracy consistently improved following fine-tuning. For example, the accuracy of Meditron-7B on the MedQA test set increased from 27.02% to 47.13%. Other models also had similar improvements. The results suggest the effectiveness of domain-specific adaptation.

**Exact- and answer option regeneration-match measures**. Table 3 presents detailed memorization results across the measures during the fine-tuning over standard benchmarks. Figure 2 further compares the answer option regeneration results. Additional results including statistical analysis are further provided in the Supplementary Material Section S3. It is important to note that the baseline models Llama 2 and Llama 3 do not publicly release their fine-tuning datasets. In contrast, LLaMA provides access to its fine-tuning data, which does not include MedQA or MedMCQA. Notably, LLaMA had significantly lower memorization across all measures compared to the other baselines. For example, on MedQA, Llama 3 has a 30-token memorization ratio of 3.10%, while LLaMA had 0.0%; for answer option regeneration, the results are 6.45% for Llama 3 vs. 2.80% for LLaMA. The difference at MedMCQA is even higher for answer option regeneration (14.22% vs. 6.04%). These results suggest that LLaMA serves as a more realistic and transparent baseline under the fine-tuning setting. In addition to LLaMA-based models, the fine-tuned gpt-oss models showed similar patterns compared to their non–fine-tuned counterparts.

Another notable observation is that the fine-tuned PMC-LLaMA model released by the original study [17] (denoted as PMC-LLaMA (Original)) shows the highest memorization ratios across nearly all evaluation measures, with significantly higher values than those of other models. For example, it achieves a 30-token memorization ratio of 18.20% on MedQA at input length $l$= 50 (up to 40% with longer input lengths, as shown in Supplementary Material Section S3), whereas other fine-tuned models range between 3.15% and 4.35%. To further validate, we independently fine-tune PMC-LLaMA using the same approach described above, denoted as PMC-LLaMA (Reproduced). The results show that the reproduced model did not demonstrate the high memorization observed in the original release. Instead, it shows memorization ratios comparable to other LLMs fine-tuned on the same datasets. One possible explanation is that the original model may have been over-fine-tuned on its training data, potentially leading to overfitting and increased memorization. Previous studies have suggested that the model generalized poorly to independent benchmarks and had hallucinations [35,36]. We manually examine the model outputs and also observe frequent repetitions and hallucinations (see Supplementary Material Section S5 for examples).

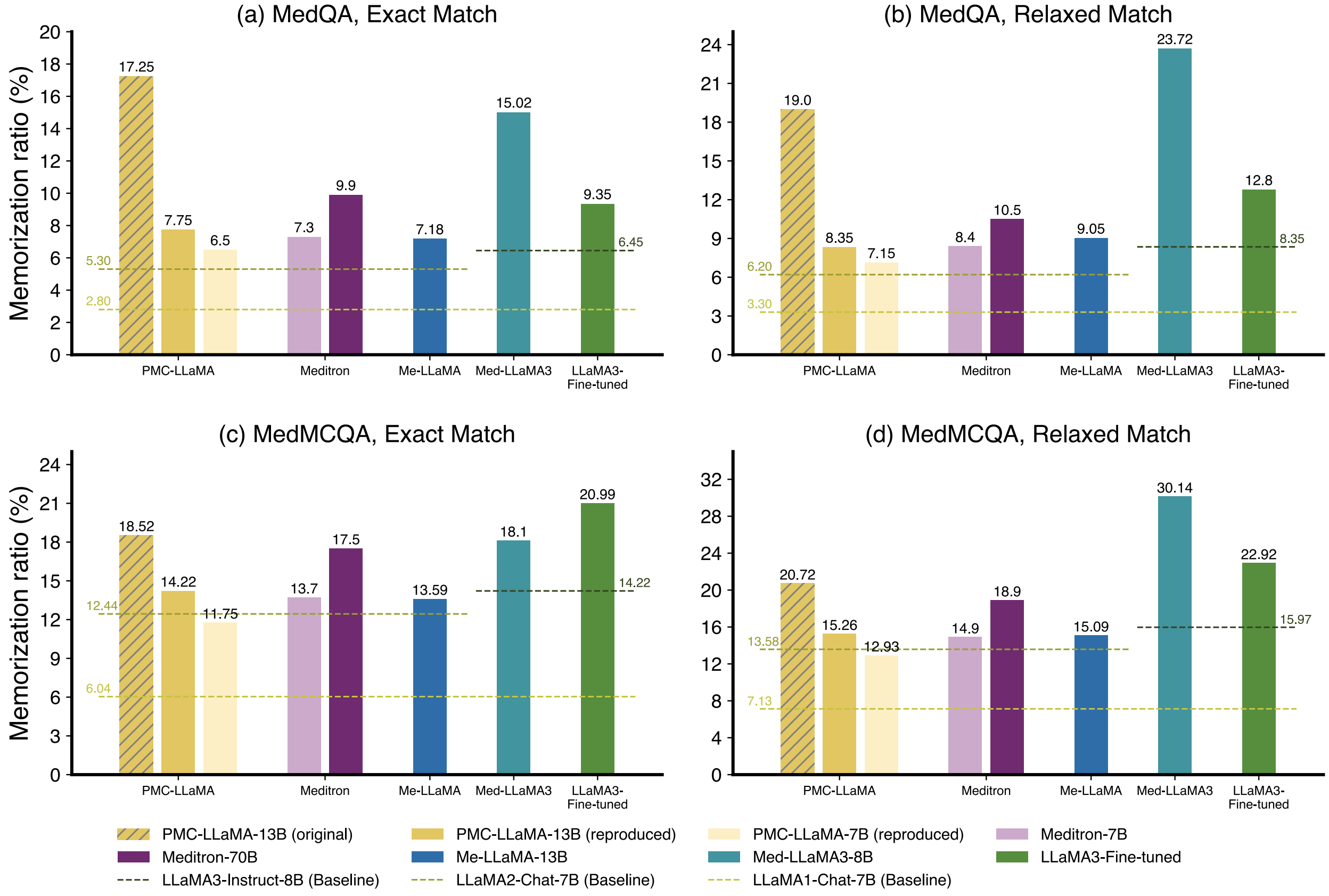

Figure 2. Results in addition to Table 3 showing large language models (LLMs) regenerating removed answer options after fine-tuning. LLMs fine-tuned on standard medical benchmarks are compared with their respective baselines.

**Comparison with baselines**. Interestingly, compared to their baselines, the fine-tuned LLMs do not show substantially higher exact memorization as observed in the continued pretraining stage. However, they demonstrate significantly higher regeneration of answer options than the baselines. For example, on MedQA, the fine-tuned Med-Llama-3 regenerates ~15% of the removed answer options (from 2,000 total evaluation instances), compared to 6.45% for Llama 3 and 2.80% for LLaMA. Similar patterns are observed for other models and datasets. We further conducted an additional analysis to assess the effect of prompt variations on memorization ratios. Specifically, we evaluated the fine-tuned Med-Llama-3 model using an alternative prompt that was semantically equivalent to the prompt used for answer option regeneration, while keeping all other evaluation settings unchanged. The detailed prompts and results are reported in Supplementary Material Section S3.4. The memorization ratios showed minimal differences across prompts on both benchmarks, with 15.02% vs. 15.06% on MedQA and 18.10% vs. 18.01% on MedMCQA.

**Comparison between continued pretraining and fine-tuning stages.** The results suggest that LLMs may show different patterns of memorization during the fine-tuning stage compared to the continued pretraining stage. For example, Med-Llama-3 regenerates over 18% of the exact answer options in MedMCQA, while its 30-token exact memorization ratio remained low at 0.19%. Similarly, Meditron-70B regenerates answer options in over 18% of MedMCQA

instances—suggesting it memorized portions of the fine-tuning dataset—yet shows only 0.25% of responses with 30-token consecutive matches. In contrast, during continued pretraining, Meditron-70B shows much higher 30-token memorization, such as over 20% on the Clinical guidelines dataset. We note two potential explanations for these differences. First, compared to continued pretraining datasets, which often consist of entire documents, fine-tuned datasets typically comprise shorter text segments, leading to variations in memorization patterns. Second, the training loss functions employed during fine-tuning differ from the regression loss used during pretraining.

**Examination of memorization on continued pretraining after fine-tuning.** We further assess whether medical foundation language models retain memorized content from their original continued pretraining datasets after undergoing fine-tuning, using Meditron as an example. Specifically, we evaluate the memorization of Meditron-7B on its original continued pretraining corpora following fine-tuning. Figure S3.3 (Supplementary Material Section S4) shows the exact memorization ratio of fine-tuned Meditron-7B on pretraining datasets, Clinical guidelines and MEDLINE abstracts. Interestingly, the results consistently show that the memorization ratios on the pretraining datasets are very similar and only slightly reduce after fine-tuning. For instance, the consecutive 50 of the fine-tuned Meditron is 8.85% on Clinical guidelines compared to 7.90% for the pretrained Meditron. Similar trends are observed for MEDLINE abstracts 0.33% vs. 0.25%.

Given the results, we further investigate whether continued pretrained and fine-tuned models memorize the same content, despite exhibiting very similar memorization ratios. Specifically, we examine all the memorized instances consisting of consecutive 50 token matches from the continued pretrained and fine-tuned Meditron models. The commonly memorized instance ratio is computed as the intersection of memorized instances divided by their union. Similarly, the uniquely memorized instance ratio is calculated as the number of instances uniquely memorized by that model divided by the union. Table 4 shows the results of datasets Clinical guidelines and MEDLINE abstracts across various input lengths. The results indicate that the fine-tuned models could memorize approximately 70% of the exact content memorized by their pretrained models. When the input length reaches 500, the overlap increases to over 87%. More results of other datasets are provided in the Supplementary Material Section S4. Additionally, fine-tuned models introduce a small proportion of new consecutive 50 token instances. Overall, these findings suggest that LLMs preserve a significant amount of memorized content from continued pretraining even after fine-tuning. Additionally, fine-tuning introduces new memorized elements—particularly task-specific content such as regenerated answer options.

Table 4. Results of whether pretrained and fine-tuned Meditron-7B models memorize the same content on the continued pretraining datasets, Clinical guidelines and MEDLINE abstracts, with input length $l$=50, 100, 200, 300, 500 tokens. Pretrained Meditron-7B unique and fine-tuned Meditron-7B unique: the uniquely memorized instance ratio is calculated as the number of instances uniquely memorized by the pretrained and fine-tuned Meditron-7B model divided by the union, respectively. Common: The commonly memorized instance ratio is computed as the intersection of instances memorized by both the pretrained and fine-tuned Meditron-7B model divided by their union.

| **Datasets** | | **$l$=50** | **$l$=100** | **$l$=200** | **$l$=300** | **$l$=500** |
|---|---|---|---|---|---|---|
| Clinical guidelines | Continued pretrained Meditron-7B unique | 18.97% | 25.31% | 16.07% | 14.95% | 6.65% |
| | Common | 71.79% | 61.73% | 66.96% | 71.89% | **87.16%** |
| | Fine-tuned Meditron-7B unique | 9.23% | 12.96% | 16.97% | 13.16% | 6.19% |
| MEDLINE abstracts | Continued pretrained Meditron-7B unique | 24.24% | 20.00% | 30.00% | 34.21% | 22.99% |
| | Common | **75.76%** | 70.00% | 55.00% | 50.00% | 59.77% |
| | Fine-tuned Meditron-7B unique | 0.00% | 10.00% | 15.00% | 15.79% | 17.24% |

**Additional approximate and semantic measures.** The additional results of approximate and semantic measures are provided in Supplementary Material Section S3.1. Similarly, these measures identify more potentially memorized instances compared to the exact measures. Qualitative examples illustrating these cases are also presented in Supplementary Material Section S3.3.

## 3.3 Memorization results of LLMs fine-tuned over clinical data

As mentioned, in addition to fine-tuning over standard medical benchmarks, we further evaluate the memorization of LLMs fine-tuned over clinical data through a case study on LLM-assisted disease diagnosis.

**Comparison of diagnostic accuracy with and without fine-tuning.** Figure 3A presents the top-$k$ diagnostic accuracy of LLMs on the test set, following the same evaluation settings established in prior work [13,37,38]. The results consistently show that both Llama-3-Instruct and Med-Llama-3 achieved improved performance after fine-tuning; and Med-Llama-3 slightly outperformed Llama-3-Instruct overall. For example, the accuracy of Llama-3-Instruct improved from 48.6% to 54.8% at top-1 ($p < 0.0001$), 63.2% to 67.7% at top-2 ($p<0.0001$), and 69.0% to 71.4% at top-3 ($p=0.00012$) after fine-tuning. Similar trends are observed for Med-Llama-3, which show consistent gains across all $k$ values. The improvements are particularly notable in specific medical specialties. Figure 3B further compares top-1 diagnostic accuracy across medical specialties with at least 100 patient records. The results show that fine-tuning led to greater improvements in specialties such as cardiology (+11.0%) and nephrology (+12.6%). These results demonstrate the value of domain adaptation in enhancing LLM performance for medical applications.

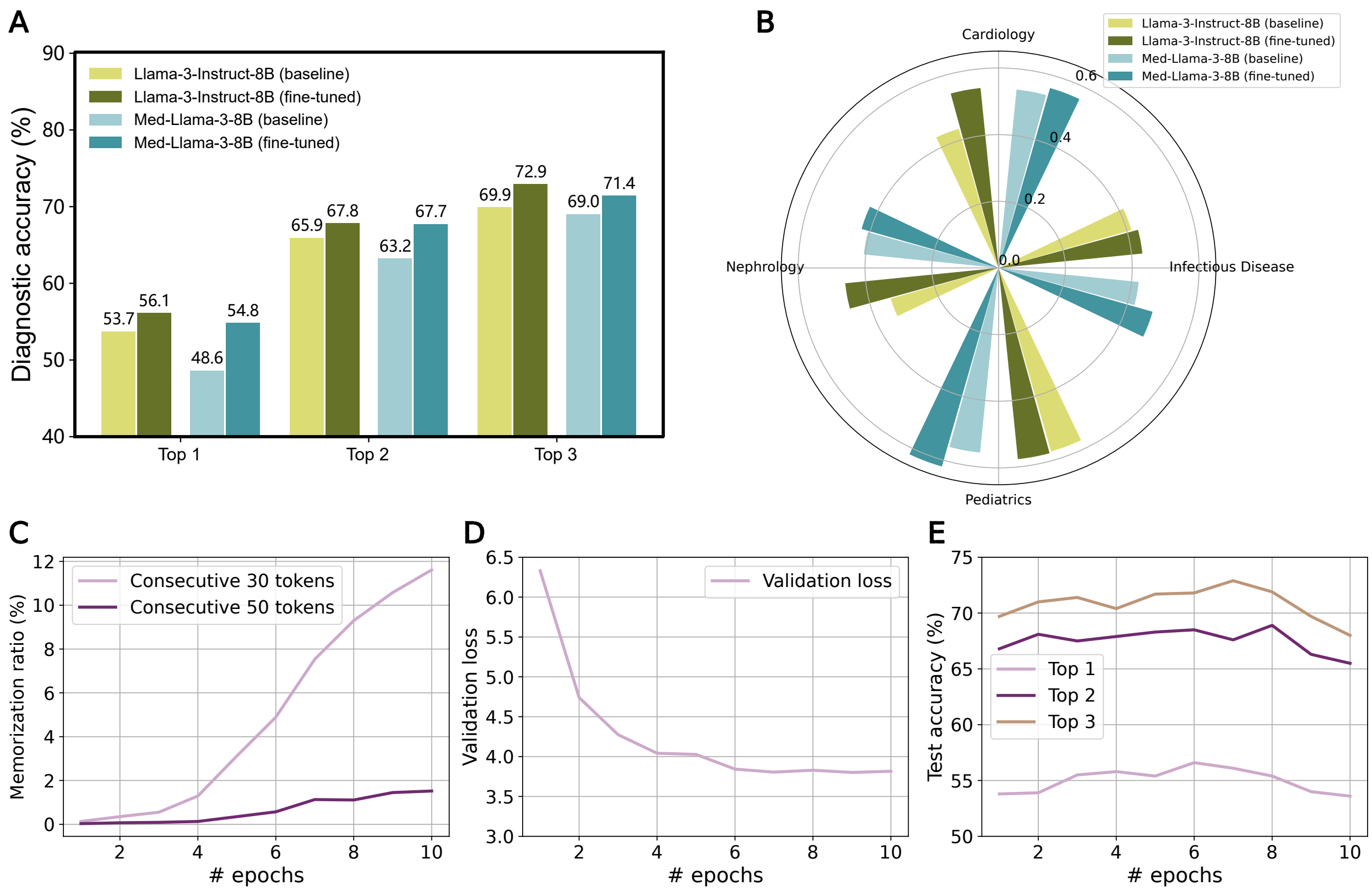

Figure 3. Diagnostic performance and training dynamics of fine-tuned medical large language models (LLMs). (A) Overall top-K diagnostic accuracy across models on the test set. Non-fine-tuned models serve as baselines for comparison. (B) Top-1 diagnostic accuracy by medical specialty on the test set. (C) Memorization ratios of consecutive 30- and 50-token spans in the training set across training epochs. (D) Cross-entropy validation loss on the validation set. (E) Top-K diagnostic accuracy (Top-1, Top-2, Top-3) on the test set. Model checkpoints were saved after each epoch to enable measurement of changes in memorization ratio (on the training set), validation loss, and test set accuracy over time. Fine-tuning details and model selection criteria are described in Section 5.3. Models were fine-tuned on the training set only, and the best checkpoint was selected based on the lowest validation loss.

Table 5. Results of 30- and 50-token memorization ratios for Llama-3-Instruct-8B and Med-Llama-3-8B models with and without fine-tuning over the training set of the clinical data across different input lengths. The fine-tuning procedure and model selection are described in Section 5.3. Models were fine-tuned on the training set only, and the best checkpoint was selected based on the lowest validation loss.

| Model | Type | Consecutive 30 tokens | | | | | Consecutive 50 tokens | | | | |
|---|---|---|---|---|---|---|---|---|---|---|---|
| | | $l$=50 | $l$=100 | $l$=200 | $l$=300 | $l$=500 | $l$=50 | $l$=100 | $l$=200 | $l$=300 | $l$=500 |
| Llama-3-Instruct-8B | Non-fine-tuned (Baseline) | 0.00% | 0.08% | 0.69% | 4.17% | 7.24% | 0.00% | 0.02% | 0.13% | 0.77% | 1.63% |
| | Fine-tuned | 2.29% | 3.89% | 17.26% | 23.82% | **28.17%** | 0.22% | 0.25% | 1.66% | 4.79% | **5.47%** |
| Med-LLaMA3-8B | Non-fine-tuned (Baseline) | 0.00% | 0.06% | 1.03% | 5.18% | 8.82% | 0.00% | 0.03% | 0.22% | 0.99% | 2.41% |
| | Fine-tuned | 4.63% | 7.53% | 22.27% | 25.88% | **30.52%** | 0.63% | 1.13% | 3.69% | 4.78% | **5.59%** |

**Exact-based memorization ratio.** Table 5 presents the 30- and 50-token memorization ratios for the LLMs with and without fine-tuning across varying input lengths. The results show a consistent increase in memorization following fine-tuning for both medical foundation language models and general-purpose LLMs. For example, the 30-token memorization ratio for Med-Llama-3 increases from 0.00% to 4.63% at input length $l$ = 50 after fine-tuning, with similar increases observed across other input lengths. A similar trend is also observed for Llama-3-Instruct.

**Changes in memorization ratio and accuracy over the course of training.** As described above, to further quantify how memorization changes in relation to model accuracy during fine-tuning, we saved model checkpoints at each epoch for up to 10 epochs. For each checkpoint, we computed the memorization ratio for consecutive 30- and 50-token spans on the training set, the cross-entropy validation loss on the validation set, and the diagnostic accuracy on the held-out test set. Figure 3C, 3D and 3E present the results for the fine-tuned Med-Llama-3. The memorization ratio increased substantially as early as epoch 3 or 4 (Figure 3C), while the validation loss continued to decrease steadily until epoch 7 (Figure 3D). Diagnostic accuracy also peaked at epoch 7 (Figure 3E). After that point—when early stopping was triggered—diagnostic accuracy declined, and memorization increased from 7.53% at epoch 7 to 11.61% at epoch 10. The trends for 50-token spans were consistent. These findings suggest that memorization could increase in the early phases of training and is not only because of overfitting due to prolonged training.

**Analysis on memorized potentially sensitive information**. As detailed in Section 5.3, we further analyze the memorization of sensitive content, including PHI based on standard definitions from prior studies [32–34], as well as other potentially sensitive information beyond PHI. For PHI, Table 6A summarizes the memorization results for the fine-tuned Llama-3-Instruct-8B model across 10,000 responses. As noted earlier, due to the scale of the data, we adopt a combined automatic and manual approach for PHI detection. A state-of-the-art PHI detection tool is applied [33], and each identified PHI instance is manually verified against the corresponding source patient notes to confirm its presence. The fine-tuned Llama-3-Instruct-8B reproduces 3,192 PHI instances, as shown in Table 6A. In addition, we also manually examine potentially sensitive content beyond the standard PHI definition. Specifically, we review 200 randomly selected outputs from Llama-3-Instruct-8B where the ROUGE-L score over 0.2 (indicating meaningful overlap between the generated and source texts). The results are summarized in Table 6B. In total, the model reproduces 128 instances of potentially sensitive information. This includes 30 PHI instances that are missed by the automated PHI detection tool, and 98 additional instances of sensitive content beyond standard PHI. These include highly sensitive diagnoses, family relationships, healthcare facility names, and treatment details. Importantly, this evaluation is conducted on patient records without deidentification in order to directly assess the extent of memorized sensitive content in LLMs. While automatic deidentification is a standard pre-processing step for clinical data [39], the results collectively suggest that applying deidentification may still not be sufficient as (1) automated deidentification tools may fail to detect all PHI, as evidenced by the 30 missed instances identified through manual review; (2) potentially sensitive information beyond PHI definition, such as those noted above, is more challenging to detect and preprocess automatically.

Table 6. Categories of memorized and potentially patient-sensitive information in model outputs. (A) Categories of memorized protected health information (PHI) detected in 10,000 outputs of the fine-tuned Llama-3-Instruct. (B) Categories of potentially patient-sensitive information identified in a random sample of 200 memorized text

samples with ROUGE-L > 0.2. PHI definitions follow previous studies [32–34]; detection used the tool [33] and all identified items were manually verified against the corresponding original patient records. Note that the original PHI is replaced with synthetic counterparts for demonstration.

A

| PHI entities | Frequency | Synthetic examples |
| --- | --- | --- |
| Dates/ ages | 2,212 | The patient is a **90-year-old** [synthetic] female presenting with shortness of breath and was examined on **09-11-2024** [synthetic]. |
| Names | 901 | No acute cardiopulmonary abnormality, reported and signed by: **Dr. John Smith** [synthetic]. |
| Locations | 79 | The patient lived on **Mira Avenue** [synthetic] and was sent to a rehab in **Brightvale Medical Center** [synthetic]. |

B

| Categories | Descriptions | Categories | Frequency | Synthetic examples |
| --- | --- | --- | --- | --- |
| PHI entities (missed by the automatic de-identification tool) | Individually identifiable health information | Genders | 23 | An 80 y.o. female presents with difficulty walking |
| | | Dates/ ages | 5 | |
| | | Locations | 2 | The patient lived on A Street |
| Highly sensitive diagnoses | Information that may cause social stigma or privacy harm | Abuse history: substance-use, alcohol use, etc | 19 | The patient has a history of ongoing substance use, alcohol use disorder, tobacco use. |
| Family relationships | Information about family members | Medical history provider | 32 | The daughter provided medical history. |
| | | Family trauma | 2 | Patient states her father's died and she cannot go on. |
| Healthcare facility | Information about hospitals or physicians | Hospital names | 5 | The patient was sent to Brightvale Medical Center |
| | | Physician names | 1 | Case discussed with Dr. Mark |
| | | Hospital locations | 3 | The patient was admitted to Bayside Medical Center. |
| Treatment details | Sensitive due to cost, rarity, or uniqueness of therapy | Laboratory findings | 29 | With history of known lumbar spinal stenosis, recent acute onset lower back pain |
| | | Medications | 7 | A 47 y.o. male on Medrol Dosepak. |

| | Description | Representatives | Recommendations |
|---|---|---|---|
| **Beneficial memorization**<br>*Recommend* | **Accurate recall of domain and task knowledge** that supports factual accuracy and reasoning in medical applications | **Biomedical concepts** (e.g., gene-disease associations)<br>**Clinical guidelines** (e.g., diagnostic criteria)<br>**Referenced literature** (e.g., for evidence attribution)<br>**Task-specific knowledge** (e.g., acquired during fine-tuning) | **Enhance during continued pretraining and fine-tuning** (e.g., incorporate domain-specific training objectives such as contrastive concept linking)<br>**Monitor preservation across adaptation** (e.g., evaluate whether domain knowledge is retained after fine-tuning new tasks) |
| **Uninformative memorization**<br>*Low to intermediate risks* | **Repetition of boilerplate or templated language** that adds little to domain or task specific knowledge | **Document disclaimer** (e.g., … is intended to be an educational tool)<br>**Statement** (e.g., redistribution or use for…)<br>**Structural headers** (e.g., introduction, methods, …)<br>**Formatting instructions** (e.g., references should follow …) | **Surface-level memorization limits deeper domain knowledge and reasoning**<br>**Select diverse and distinct training examples for continued pretraining** (e.g., through data deduplication or clustering)<br>**Encourage reasoning-focused learning other than rote memorization** (e.g., incorporating reasoning-based post-training) |
| **Harmful memorization**<br>*High risks* | **Regeneration of data-specific content or sensitive information** that reduces generalization, violates privacy policies, and poses risks to downstream use | **Sensitive patient information** (e.g., Protected Health Information and other sensitive content beyond standard identifiers)<br>**Verbatim clinical narratives** (e.g., exact replication of descriptions, imaging impressions, or clinical notes from training data)<br>**Dataset artifacts** (e.g., reproduced answer options from medical QA benchmarks) | **Incorporate strategies to penalize dataset-specific memorization during training** (e.g., explore adversarial learning related approaches to discourage reliance on training artifacts)<br>**Standard processing of clinical data such as automatic de-identification is necessary but may not be sufficient** (e.g., consider privacy-preserving techniques) |

**General recommendations**

1. **Evaluate memorization alongside accuracy-based metrics.** To support reporting standards and downstream accountability, we recommend reporting both exact memorization metrics and task-specific measures at a minimum.
2. **Monitor memorization during task adaptation.** Memorization persists in LLMs in the medical domain. We suggest evaluating both retained memorized knowledge and generalization to prior tasks.
3. **Prioritize checking top tokens in model outputs.** Memorization often occurs in top tokens of LLM responses. We recommend starting evaluations there.

Figure 4. Types of memorization in large language models (LLMs) in medicine, representative examples, and recommended strategies.

# 4. Discussion

## LLMs demonstrate prevalent memorization when adapted to medicine

To adapt LLMs for medicine, researchers may build medical foundation language models by continuing pretraining on medical corpora followed by fine-tuning [17,18], or by directly fine-tuning general-domain LLMs on labeled medical data [21]. The results show that memorization of training data is prevalent across both approaches. For example, during continued pretraining, models have a up to 22% 30-token memorization ratio on specific corpora at input length $l$ = 50 (Table 2); during fine-tuning on standard medical benchmarks, models regenerate up to 18% of removed answer options (Table 3); similarly, during fine-tuning on real-world clinical data, models have a ~5% 30-token memorization ratio at $l$ = 50 and over 3,000 memorized PHI instances across 10,000 patient training samples (Tables 5 and 6A). In addition to strict exact measures, additional metrics including approximate and semantic measures also show consistent memorization ratios and suggest a higher prevalence of memorized instances than exact measures alone (e.g., Tables 2 and S2.3). Our report of prevalence of memorization in the medical domain may serve as a benchmark for future work. The memorization ratios we observed were significantly higher than previously reported for LLMs in the general domain—prior work found memorization ratios of only ~0.1% using exact measures [27,40]. We anticipate that a key underlying reason for this higher memorization prevalence is the current continued pretraining or fine-tuning approaches are mostly adapted from the general domain with minimal methodological changes. For example, continued pretraining of medical foundation language models typically involve simply continuing the pretraining of general-domain LLMs on medical corpora without additional methodological adaptations [17,18]. While straightforward, this approach likely leads to more prevalent memorization. General-domain LLMs are typically pretrained on heterogeneous datasets such as web pages and news articles, offering diverse content and stylistic variety [41]. In contrast, medical domain-specific LLMs are pretrained on more homogeneous medical corpora, which are closely related and uniform in both content and style. The same applies to the fine-tuning process. Given the limited curated datasets available in the medical domain, studies often fine-tune LLMs by combining closely related medical datasets (e.g., combining several medical QA datasets for instruction tuning [17–19]), which likely further contributes to the observed high prevalence of memorization.

**LLMs demonstrate distinct and unique memorization characteristics**

Figure 4 illustrates the different types of memorization observed when adapting LLMs to the medical domain through continued pretraining or fine-tuning and provide representative examples based on the evaluation results. First, LLMs are capable of accurately recalling domain-specific knowledge, such as biomedical concepts and clinical guidelines (e.g., Table S2.2). Second, they can also regenerate templated language, including standard disclaimers and repetitive statements (also shown in Table S2.2). Third, LLMs can memorize data-specific content and potentially sensitive clinical information. For example, LLMs reproduce removed answer options when fine-tuned on medical QA benchmarks (e.g., Figure S2.2), regenerate PHI when fine-tuned on clinical data (e.g., Table 6A), and regenerate other potentially sensitive information beyond standard PHI definitions (e.g., Table 6B).

In addition, a notable and unique observation is that fine-tuning does not substantially eliminate memorization acquired during the continued pretraining stage in the medical domain. Instead, memorization of the continued pretraining corpora persists, with models retaining a substantial portion of previously memorized content even after fine-tuning. For example, the results show that up to 87% of memorized content from the continued pretraining stage remains unchanged following fine-tuning (Table 4).

Another observation from the fine-tuning results on clinical data (Figure 3C, 3D, 3E), where we evaluated memorization ratio, validation loss, and test set accuracy at each epoch, is that memorization can increase early in training, while the validation loss is still decreasing. This suggests that memorization does not necessarily emerge only in later epochs due to overtraining. Collectively, those memorization characteristics differ markedly from what has been reported in the general-domain literature [27,40].

The study also reports several factors that systematically influence memorization. Larger models exhibit higher memorization ratios (e.g., Meditron-70B vs. Meditron-7B). Longer inputs lead to increased memorization across different input length $l$. More fine-tuning epochs are associated with higher memorization ratios, and models exposed to the target dataset during training exhibit significantly higher memorization ratios. The training stage also matters: continued pretraining tends to produce more verbatim memorization, whereas fine-tuning produces more task-specific memorization patterns. In contrast, common decoding parameters such as temperature and top-$k$ do not result in significant changes to memorization.

**The memorization of LLMs impact both method development and adoption**

Memorization has both positive and negative implications, which impact the development and adoption of LLMs in medicine. Figure 4 illustrates these potential impacts.

On the positive side, the results show that continued pretraining enables LLMs to recall clinical guidelines, biomedical concepts, and medical literature (Table S2.2); fine-tuning over medical data also enables LLMs to more effectively follow task-specific instructions for medical applications, such as improving diagnostic accuracy (Figure 3A, 3B). This type of memorization can be beneficial, as it may enhance factual accuracy and reasoning in downstream medical tasks.

However, the negative implications of memorization are substantial and two-fold. First, the results show that LLMs frequently memorize templated language, such as document disclaimers and formatting (Table S2.2). This form of uninformative memorization largely reflects surface-level pattern matching rather than any type of domain reasoning. As a result, this type of memorization may not contribute to downstream task performance and can lead to overfitting and poor generalization. For instance, one fine-tuned LLM released by the original study has a 30-token memorization ratio of over 18% at $l$ = 50 and over 40% with longer input lengths (Figure 2; Supplementary Materials Section S3) on its fine-tuned medical datasets. It fails to follow other task instructions and generates repetitive or hallucinated content (Supplementary Materials Section S5), consistent with the findings reported in previous studies [15,18]. These results underscore the need for improved training strategies that move beyond surface-level memorization to promote the acquisition of domain knowledge and generalization capabilities. Recent work on post-training for reasoning offers a potential direction to revisit current strategies for adapting LLMs to medicine [42,43].

Second, LLMs can also regenerate dataset-specific information or sensitive clinical content. This form of memorization may be harmful and pose risks in practice. In this study, we perform a case study on fine-tuning LLMs on real-world clinical data for assisting disease diagnosis. The results demonstrate that fine tuning is effective and essential for medical applications. For example, the results show consistent improvements in overall top-$k$ diagnostic accuracy and gains of over 10% in specific medical specialties for both medical foundation models and general-purpose LLMs (Figure 3A, 3B). However, it also shows potential concerns on memorizing sensitive clinical data. For instance, LLMs fine-tuned on clinical data have a ~5% 30-token memorization ratio at $l$ = 50 (Table 5) and reproduce over 3,000 PHI instances across 10,000 patient training samples (Table 6A). As mentioned, while de-identification is a common pre-processing step in clinical data workflows, the findings suggest it may not be sufficient. Manual review of 200 outputs reveals 30 PHI instances missed by the automated de-identification tool, as well as ~100 instances of potentially sensitive information—such as highly sensitive diagnoses, healthcare facility names, and family relationships—which are beyond the standard PHI definition and are more difficult to detect automatically (Table 6B). This suggests that practitioners could consider evaluating memorization in addition to accuracy-based metrics during model development and deployment. Indeed, recent reviews also highlight the gap in current evaluations reporting of LLMs, which has largely focused on accuracy on standard benchmarks [44–46]. In contrast, our study presents a downstream case study that examines both the benefits and risks of fine-tuning LLMs on real-world clinical data. The findings also motivate future research to further examine the impact of memorization on the safe and responsible deployment of LLMs in downstream medical applications.

**Recommendations and call for community efforts**

The memorization evaluation pipeline covers key adaptation stages and has been applied to both general-domain and medical LLMs. The results for LLaMA-based models, as well as more recent models such as gpt-oss, show similar patterns (see Tables 3 and S2.4). We have made the pipeline publicly available, allowing the community to directly apply it to emerging models and datasets. We provide practical recommendations based on the findings. Figure 4 outlines detailed suggestions by memorization type and general recommendations on managing memorization. The goals are to facilitate beneficial memorization such that LLMs capture domain-specific knowledge and reasoning for domain adaptation and preserve it when fine-tuning on new tasks; minimize uninformative memorization so that models learn deeper knowledge rather than surface-level patterns; and mitigate harmful memorization to avoid reproducing data-specific information or sensitive clinical content.

Figure 4 provides detailed suggestions per memorization type. For instance, for uninformative memorization, revisiting the current approach from both data and method perspectives could offer potential solutions. From a data perspective, we suggest increasing dataset depth and breadth during domain adaptation. This can help reduce repetitive patterns and improve the model's ability to generalize beyond surface-level patterns (e.g., Tables 2 and S2.2). Techniques such as data deduplication or clustering to select distinct representatives can help [47]. Rejection sampling (where low-quality or redundant samples are filtered out) or the use of quality indicators may further enhance training data quality [48]. From a method perspective, we recommend exploring recent advances in reasoning-based post-training. Pilot studies have shown that post-training on reasoning tasks could improve generalization and capture domain knowledge [42,43], potentially helping models move beyond surface-level memorization. For harmful memorization where LLMs may regenerate data-specific content or sensitive patient information, several mitigation strategies have been explored. In addition to standard de-identification techniques and the use of early stopping to prevent overfitting (as demonstrated in Figure 3C, 3D, 3E), which effectively reduce memorization, recent studies are investigating privacy-preserving methods for LLMs [49–51], such as applying differential privacy during fine-tuning on sensitive datasets [51]. Another parallel direction involves using retrieval-augmented generation (RAG) to equip LLMs with domain-specific knowledge without retraining the models themselves [52–54]. Both are active areas of research for further exploration.

In addition, we call for community efforts to strengthen reporting guidelines for LLMs in medical applications [55,56]. Most studies focus on reporting accuracy-based metrics for LLMs in medical applications [57,58]. We argue that memorization should be evaluated alongside accuracy-based metrics as downstream accountability. At a minimum, we suggest reporting exact measures and task-specific memorization measures. Also, to date, integrating LLMs into healthcare remains an open challenge [59]. Pioneering studies have explored potential deployment pathways and their trade-offs; e.g., the use of third-party APIs, where privacy, security, and regulatory concerns persist [59–61]. Given that LLM memorization, particularly when models are trained on real-world clinical data, poses potential risks such as the regeneration of sensitive clinical information, we encourage future deployment frameworks to account for memorization-related vulnerabilities as part of broader safety and compliance protocols.

**Limitations and future work**

This study has several limitations. First, this study does not include targeted testing to assess the maximal extent of memorization that can occur when adapting LLMs for medical applications. As a preliminary effort, we focused on three major adaptation scenarios: continued pretraining, fine-tuning on standard benchmarks, and fine-tuning on clinical data. Where possible, we used the exact models and datasets released by the original studies and followed established training and evaluation practices to reflect realistic usage scenarios and inform both developers and users of LLMs in medicine. Future work is needed to explore worst-case behaviors, such as simulating adversarial attacks designed to extract sensitive information [62]. This line of research, consisting of both offensive and defensive strategies for LLMs, is beyond the scope of our current study but has been actively explored in the general-domain literature [63,64].

Second, while we examined several factors that may influence memorization, such as model size, training stage, and input length, and included both medical foundation models and general-purpose LLM variants across different adaptation scenarios, many additional dimensions remain to be explored. For example, different prompting strategies could impact memorization. In this study, we tested only two prompt variations with equivalent semantic meaning and observed consistent memorization ratio (see Supplementary Material Section 3.4). Other downstream tasks, such as clinical information extraction and medical report summarization, as well as the evaluation of additional LLM families, also represent important directions for future investigations.

Third, because different LLMs are continued pretrained or fine-tuned on different datasets, and due to varying availability (e.g., some fine-tuned models are not released), we have to evaluate different models on different datasets and reproduce some models when they were not publicly available. In such cases, we provide the code and reproduced models via our repository. As a result, direct inter-model comparisons are not warranted. Instead, comparisons should be made between each model and its corresponding baseline. This limitation underscores the need for standardized benchmarks to support more consistent and comparable evaluations of memorization.

Fourth, while our study focuses on memorization of LLMs in medicine, it is also important to explore the broader interplay between memorization, overfitting, and generalization of LLMs in medical applications. In this study, we conducted a preliminary analysis (Figure 3C, 3D, 3E), which suggests that memorization does not necessarily emerge only in the later stages of training due to overfitting. Notably, recent work on *grokking* [65,66] describes a phase transition wherein models initially memorize the training data and only begin to generalize to the test set after prolonged training. Our preliminary results are related but also distinct: memorization appears to emerge early in training, even as validation loss continues to improve. Further investigation is needed to understand these dynamics more systematically and assess their implications on LLMs in medicine.

In addition, we encourage future work to explore the memorization of LLMs in medicine in other practical settings, including but not limiting to approaches that call closed-source LLMs via third-party APIs when the training data cannot be accessed [67]. Having access to memorized or regenerated training data reproduced in LLM outputs can constitute a privacy risk. For instance, regressive sampling that simulates possible training samples and then prompts LLMs to regenerate data snippets may be explored [27].

Finally, as discussed, memorization in LLMs has both positive and negative implications. Moving forward, it will be important to better understand these trade-offs and identify contexts in which different levels or types of memorization are appropriate. Collectively, these directions will contribute to a more comprehensive understanding of memorization and support the development of effective solutions for the safe and responsible adoption of LLMs in medicine.

# 5. Methods

As noted above, to adapt LLMs for medicine, studies may build medical foundation language models (through continued pretraining followed by supervised fine-tuning) or directly fine-tune LLMs using labeled medical datasets. In this study, we systematically assess memorization behavior across both the continued pretraining and fine-tuning stages. For fine-tuning, we also evaluate both scenarios where studies fine-tune LLMs using standard medical benchmarks (e.g., medical question answering) or fine-tune using clinical data (e.g., clinical notes). The case study in Section 5.3 is approved by the Yale Institutional Review Board (IRB) under protocol number 2000037687. Table 1 provides an overview of the models and datasets. Detailed descriptions are provided below.

## 5.1 Evaluation of LLMs memorization originating from continued pretraining

Continued pretraining further adapts LLMs into medical foundation language models by training them on large-scale medical corpora (such as clinical notes and biomedical literature) using self-supervised learning objectives, such as next-token prediction [17].

**Models and datasets.** As demonstrated in Table 1, we evaluate six continued-pretrained versions of four medical foundation language models: PMC-LLaMA (7B and 13B) [17], Meditron (7B and 70B) [18], Me-LLaMA (13B) [19], and Med-Llama-3 (8B) [22]. These models are selected for two primary reasons: (1) they have been widely used in the medical domain as representative medical foundation language models [68], and (2) both the model weights and their continued pretraining datasets are publicly available, enabling us to compare generated outputs against their respective training corpora to quantify memorization.

PMC-LLaMA [17] is one of the first continued-pretrained medical foundation language models. It uses Llama 2 as the backbone and further pretrained on 4.8M PubMed Central (PMC) full-text articles, with over 75B tokens. These biomedical research articles are curated from the Semantic Scholar Open Research Corpus (S2ORC) [69], based on the availability of corresponding

PMC IDs. We assess the 7B- and 13B-parameter versions of PMC-LLaMA on the dataset PMC full-text dataset.

Meditron [18], in contrast, extends its continued pretraining on a collection of mixed medical and general-domain corpora: (1) Clinical guidelines (46,000 clinical practice guidelines sourced from 16 global healthcare organizations), (2) MEDLINE abstracts (16.1 million MEDLINE abstracts), (3) PMC full-text articles (4.9 million PMC full-text articles), and (4) Experience replay (a general-domain dataset containing 494,000 documents sourced from the RedPajama dataset [20]). We evaluate the 7B- and 70B-parameter versions of Meditron on its four pretraining datasets.

Me-LLaMA [19] is developed through continued pretraining of Llama 2 using a diverse medical dataset comprising 129 billion tokens. Specifically, it is one of the first LLMs to use clinical notes for pretraining. Its pretraining dataset includes clinical notes from MIMIC-III [70], MIMIC-IV [71], and MIMIC-CXR [72], as well as biomedical literature from MEDLINE abstracts and PMC full-text articles [73]. We evaluate the 13B-parameter version of Me-LLaMA on its three pretraining datasets: MEDLINE abstracts, PMC full-text articles, and MIMIC-III.

Med-Llama-3 [22] is a recent medical foundation language model built through continued pretraining of Llama 3, extending the Me-LLaMA framework [19]. We evaluate the 8B-parameter version of Med-Llama-3 on two of its primary pretraining subsets: Clinical guidelines and MEDLINE abstracts.

**Evaluation measures.** As shown in Table 1, the models were continued-pretrained on different corpora. We evaluate memorization on the specific corpora used for continued pretraining for each model. Specifically, we quantify the memorization of LLMs by prompting the models to generate the remaining text given an input text from their associated continued pretraining datasets and comparing the generated text with the original text. From each dataset of an LLM, we divide an input text to a length-$l$ prefix and a suffix. For each input text, we prompt the model with the prefix and allow the model to generate 500-token continuations, which we then compare to the suffix. An example is provided in the Supplementary Material Section S1.1. We systematically assess using prefixes of varying lengths: $l$ = 50, 100, 200, 300, and 500 tokens. This setting is used in the studies to cover different scenarios [27,74]. A short prefix length of 50 tokens is feasible in practice where short input segments of datasets can be easily or directly obtained. Additionally, some strategies such as autoregressive sampling [75], can be designed to generate prefixes of this length to extract training data [27]. In contrast, a long prefix length of 500 tokens is employed to evaluate worst-case scenarios involving longer text snippets from the datasets [74]. To generate reproducible outputs with minimal variations, we set the temperature to 0 and employ a greedy decoding strategy to generate tokens that was used in some existing studies [75–78]. We also perform additional analysis on the effect of sampling temperature (0.1, 0.3, 0.5, 0.7, and 1.0) and top-$k$ values (10 and 50) on memorization. These values are adopted based on existing studies on decoding strategies [79]. We evaluate six measures across three categories (exact, approximate, and semantic) as detailed below.

**Exact measures.** These measures quantify the number of consecutive exact tokens that LLMs can regenerate from pretraining sets. We employ exact measures as the main evaluation measures. They are also the primary evaluation measures of memorization for LLMs in the general domain [74,80]. Specifically, we detect whether the generated text contains exact identical token sequences of length at least $\tau$ compared to the original text. We assess exact measures with $\tau$ of 30 and 50 tokens (consecutive 30 and 50 tokens, respectively) consistent to the studies in the general domain where the value of $\tau$ is empirically set to be large enough to prevent accidental overlaps [74,80]. We report the memorization ratio as the percentage of instances in the pretraining datasets where LLMs generated consecutive identical sequences of 30 or 50 tokens compared to the original instance.

**Approximate- and semantic- measures.** While exact measures are commonly used, they (1) constrain the positions of tokens without allowing for variation, and (2) assess the exact tokens rather than capturing their semantic meanings. Therefore, we also adopt complementary measures to complement. Approximate-match measures focus on substring overlaps. We use the BLEU score and the ROUGE-L score [81] between the generated tokens and ground truth tokens. These metrics are also extensively used for text generation tasks [82]. In addition, semantic-match measures focus on semantic similarity between the generated text and original text [76]. We used BERT score [82] and BART score [83]—two commonly used metrics for text semantic similarity—for assessing semantic memorization. These metrics are also widely used for text similarity tasks [82]. The calculations for both approximate- and semantic-match measures follow established approaches [84] [85]. Notably, during the continued pretraining stage, we observe prevalent repetitions and hallucinations in the responses of LLMs (examples are provided in the Supplementary Material Section S5). To mitigate potential biases, we focus the evaluation of approximate- and semantic-match measures on the first 100 tokens of the generated and original texts rather than the entire output.

**Baselines**. To establish baseline performance, we compare each continued-pretrained medical foundation language model with its corresponding general-purpose backbone model: for example, PMC-LLaMA, Meditron, and Me-LLaMA are compared with Llama 2, whereas Med-Llama-3 is compared with Llama 3. It is also important to note that the pretraining datasets for Llama 2 and Llama 3 are not publicly available, making it unclear whether they overlap with the medical corpora used in continued pretraining. To provide an additional point of reference, we also include LLaMA as a baseline. Its pretraining corpus is publicly available and does not overlap with the datasets used in the continued-pretraining of the models above.

**Bootstrapping and statistical analysis.** Given the data scale, it is computationally intensive to measure memorizations for all models across all datasets. For each dataset, we perform bootstrapping with a sample size of 100 randomly selected instances, repeating the process 100 times, and report the estimated distributions within 95% confidence intervals. We conduct a two-tailed Wilcoxon rank-sum test to quantify the statistical significance of the observed differences. The bootstrapping and statistical analysis follow existing studies [86–88]. Implementation details are summarized in Supplementary Material Section S2.2, and we also make the codes available.

## 5.2 Evaluation of LLMs memorization originating from fine-tuning over standard medical benchmarks

In contrast to continued pretraining, the fine-tuning stage involves supervised learning to optimize LLMs for specific tasks, where each task is provided with labelled input-output pairs [17,18]. A common fine-tuning approach adopted in the medical domain is instruction tuning, where each instance includes instructions describing the tasks, guidelines, and input-output pairs [17,18]. In this section, we assess memorization during fine-tuning on standard medical benchmarks, as summarized in Table 1.

**Models and datasets.** Prior studies fine-tune either medical foundation language models or general-purpose LLMs. In this evaluation, we consider both cases. Specifically, we fine-tune the medical foundation language models introduced above—PMC-LLaMA (7B and 13B), Meditron (7B and 70B), Me-LLaMA (13B), and Med-Llama-3 (8B)—on standard medical QA benchmarks. For general-purpose LLMs, we similarly fine-tune Llama 3 on the same benchmarks. In addition to LLaMA-based models, we also included gpt-oss-20B, a more recent LLM released in August 2025 [23]. Gpt-oss-20B is a reasoning-focused model based on a mixture-of-experts Transformer architecture, trained using both distillation and reinforcement learning. We compared the model with and without fine-tuning over medical QA benchmarks. Medical QA assesses the effectiveness of LLMs to answer medical questions [89] and has been widely used in prior work [10,18]. These medical foundation language models also reported fine-tuning performance on three shared medical QA benchmarks and released their fine-tuned versions in their respective original studies: MedQA [90], MedMCQA [91], and PubMedQA [90]. Both MedQA and MedMCQA are composed of MCQs, whereas PubMedQA consists of yes/no/maybe questions without predefined answer options. We use MedQA and MedMCQA for evaluation in this study. MedQA [90] consists of US Medical License Exam (USMLE)-style questions that integrate various medical knowledge (e.g., patient profiles, disease symptoms) and provide four possible options, including human-written explanatory answers. MedMCQA [91] comprises 194K four-option multiple-choice questions and contains 183K samples with explanatory answers covering 2.4K healthcare topics and 21 medical subjects. Finally, we also include an evaluation of the Llama 3 model fine-tuned on these same medical QA benchmarks. Similarly to the continued pretraining stage, we use the fine-tuned versions of those foundation medical language models made available by their respective studies for reproducibility. Specifically, PMC-LLaMA and Me-LLaMA have publicly released their fine-tuned models on MedQA and MedMCQA. For Meditron, Llama-3 and gpt-oss-20B whose fine-tuned versions are not publicly available, we fine-tune the models on the training sets of MedQA and MedMCQA following the settings utilized by the original study [18]. Specifically, we use 4-bit quantized models with Low-Rank Adaptation (LoRA, $r$=16, $\alpha$=64, dropout=0.0) and focus on tuning linear layers. The fine-tuning is conducted using the AdamW optimizer for 3 epochs, with a learning rate of 2e-5, a weight decay of 0.1, a warmup ratio of 0.01, and batch size of 16. Similarly, we fine-tune Llama 3 and gpt-oss-20B on these benchmarks using the same settings. We also make these fine-tuned models publicly available.

**Evaluation measures**. For the primary evaluation measures, we use (1) exact measures, including consecutive 30 tokens and consecutive 50 tokens, and (2) answer option regeneration measures, including exact answer option match and relaxed answer option match. The exact measures are described in detail above. Using the same approach, we prompt the models to generate the remaining text given an input text in the fine-tuned training datasets and compare the generated text with the original text. In contrast, the answer option regeneration measures are specifically designed to quantify the memorization of LLMs in Medical QA tasks. We randomly remove one answer option from each QA pair in the fine-tuned training datasets and prompt the model to regenerate the removed option. For example, the prompt is "You are a doctor, kindly generate one other option based on the patient's description and the other three options". A concrete example of the prompt, input and model output is provided in the Supplementary Material Section S1.2. For each generated answer option, we calculate two metrics: (a) Exact Answer Option Match, which indicates that the generated option matches the removed option exactly, and (b) Relaxed Answer Option Match, which indicates that the generated option contains the removed option as a substring. Additionally, we incorporate approximate and semantic measures introduced in the continued pretraining stage, as additional evaluation metrics. In addition to quantifying memorization on the training set, we also report the accuracy of the fine-tuned models on held-out datasets for both MedQA and MedMCQA. For MedQA, we report accuracy on its designated test set. For MedMCQA, since the test set is not publicly available, we follow prior studies [17,18] and report accuracy on the validation set.

**Baselines and statistical analysis**. Consistent with the continued pretraining strategy, we compare each fine-tuned model with its corresponding fine-tuned general-purpose LLM baseline. Specifically, the fine-tuned versions of PMC-LLaMA, Meditron, and Me-LLaMA are compared with Llama-2-Chat [92], while both Med-Llama-3 and fine-tuned Llama 3 are compared with Llama-3-Instruct [93]. We also include LLaMA-Chat [94] as an additional baseline, as the fine-tuning datasets for the other baseline models are not publicly available. For gpt-oss, we compared the model with and without fine-tuning. We adopt the same bootstrapping and Wilcoxon rank-sum test using a random sample size of 100 instances and repeating the process 100 times, as described in Section 5.1.

## 5.3 Evaluation of LLMs memorization originating from fine-tuning over clinical data

In addition to standard benchmarks, we further evaluate memorization during fine-tuning over clinical notes through a case study on LLM-assisted disease diagnosis. Disease diagnosis is a key AI application in the medical domain [95,96], and prior studies have adopted LLMs for disease diagnosis and reasoning [20,26]. Following a similar setup, we fine-tune LLMs on real-world patient records and assess the memorization described below.

**Dataset curation.** We randomly sample 13,000 unique inpatient records who visited the emergency department (ED) of Yale New Haven Health System and were transferred to the hospital from 2014 to 2024. LLM-assisted diagnosis has the potential to aid clinicians in diagnosing these cases [13,37]. For each patient, we curate their most recent record. The input includes the history of present illness (HPI), the corresponding HPI date, and relevant laboratory results. The output consists of a list of diagnoses, which serves as the gold standard for evaluation. The dataset is then randomly divided into three subsets: a training dataset consisting of 10,000 records, a validation dataset consisting of 2,000 records, and a test dataset consisting of 1,000 records.

**Models and baselines**. As shown in Table 1, we fine-tune both the medical foundation language model Med-Llama-3 and the general-purpose LLMs Llama 3 as representative models via instruction tuning (referred to as fine-tuned models). For example, the prompt is "You are a helpful physician assistant. Your task is to give a list of correct diagnosis based on the HPI description and other medical information of the patient". An example of an instruction, input and model output is provided in Supplementary Material Section S1.3. We fine-tune the models on the training set and selected the best model based on validation loss. Specifically, we use 4-bit quantized models with LoRA ($r$=64, $\alpha$=128, dropout=0.05) and focus on tuning linear layers. The instruction tuning is performed using the AdamW optimizer, with a learning rate of 1e-6, weight decay of 0.1, a batch size of 16. We apply the early stopping strategy [97] based on the validation loss: training is stopped after two consecutive epochs without improvement in the validation loss (early_stopping_patience=2), and the model checkpoint with the lowest validation loss is selected. Note that the test set is not used for early stopping or any model selection. The same models without fine-tuning on medical data (referred to as non-fine-tuned models) are used as baselines for comparison.

**Evaluations**. We evaluate the memorization of the fine-tuned LLMs using two complementary approaches. First, we employ exact measures including consecutive 30 tokens and consecutive 50 tokens outlined in Section 5.1. In addition, we further examine the memorization of sensitive information, including both PHI [98] and other potentially sensitive content beyond standard PHI definitions. For PHI, we adopt the standard definition from existing studies [32–34], which is based on: (1) the 18 identifiers defined under the Health Insurance Portability and Accountability Act (HIPAA, 45 CFR 164.514), covering individuals, their relatives, employers, or household members; and (2) extensions proposed in Stubbs *et. al* [99], which include information that could indirectly identify individuals, such as hospital names, clinician names, and patient professions. Given the size of the training set (10,000 instances), exhaustive manual review is infeasible. We therefore use a publicly available PHI detection model that achieves an F1-score above 95.60% on a benchmark dataset using the same PHI categories [33]. For the 10,000 training instances, we apply the PHI detector to the corresponding LLM-generated outputs, and manually verify each identified PHI by cross-referencing it with the original patient records. To capture other potentially sensitive content beyond standard PHI definitions, we also manually review a set of 200 LLM responses and categorize additional types of potentially sensitive information. Further, we also report top-$k$ diagnostic accuracy on the test set, following prior studies [13,37,38], to evaluate the effectiveness of LLMs in disease diagnosis. Finally, while we employed early stopping to prevent overfitting, we also saved model checkpoints at each epoch to further quantify changes in memorization ratio and accuracy over the course of training. At each checkpoint, we evaluated the corresponding memorization ratio, validation loss, and diagnostic accuracy.

# Data availability

The data that do not contain patient data are publicly available via https://github.com/Yale-BIDS-Chen-Lab/llm_memorization and has been archived in Zenodo https://doi.org/10.5281/zenodo.18906680 [100]. Source data are provided with this paper.

# Code availability

The codes are publicly available via https://github.com/Yale-BIDS-Chen-Lab/llm_memorization and has been archived in Zenodo https://doi.org/10.5281/zenodo.18906680.

## Acknowledgments

This study is supported by the National Institutes of Health grant 1R01LM014604 and the Intramural Research Program of the National Library of Medicine (NLM).

## Author Contributions Statement

A.L. and Q.C. designed the research. A.L., M.D., Y.Y., Y.H., Z.S., Y.F., H.K., L.Q., E.S., X.A., Q.X., R.Z., J.H., Y.F.Y., S.L., Y.C.T., L.O.M, H.C, Z.L, H.X. and Q.C. performed experiments and data analysis. A.L., H.C, H.X. and Q.C wrote and edited the manuscript. All authors contributed to discussion and manuscript preparation.

## Competing Interests Statement

None declared.

Table 1. An overview of models and datasets. Primary large language models (LLMs) and their direct baselines (i.e., models with the same backbone, such as Med-Llama-3 vs. Llama 3) are provided. Because the training corpora for baseline models (e.g., Llama 2, Llama 3) are not publicly available, we include LLaMA as an additional reference baseline; its training data is publicly accessible and does not overlap with the corpora used in continued pretraining or fine-tuning. For the continued pretraining stage, each model is evaluated on the specific corpora used for its continued pretraining. For example, Meditron is assessed on its four corresponding datasets. Note that some models use different subsets from same sources, e.g., both PMC-LLaMA and Meditron use PubMed Central (PMC) full-text articles but with different selections. Subscripts are used to denote these distinctions. For the fine-tuning stages, we fine-tune both medical foundation models and general-purpose LLMs and compare them to the non-fine-tuned counterparts.

| Primary LLMs | Model Sizes | Datasets | Number of Samples | Number of Tokens | Direct baselines | Additional baselines |
|---|---|---|---|---|---|---|
| **Continued pretraining** | | | | | | |
| PMC-LLaMA | 7B/ 13B | PMC full-text articles$_1$ | 4.8M | 75B | Llama 2 | LLaMA |
| Meditron | 7B/ 70B | Clinical guidelines | 38K | 117M | | |
| | | MEDLINE abstracts | 15.7M | 5.48B | | |
| | | PMC full-text articles$_2$ | 4.9M | 40.7B | | |
| | | Experience replay | 494K | 420M | | |
| Me-LLaMA | 13B | MEDLINE abstracts | 15.7M | 5.48B | | |
| | | PMC full-text articles$_3$ | 2.57M | 21.3B | | |
| | | MIMIC-III | 2.08M | 1.38B | | |
| Med-Llama-3 | 8B | Clinical guidelines | 38K | 117M | Llama 3 | |
| | | MEDLINE abstracts | 15.7M | 5.48B | | |
| **Fine-tuned over benchmarks** | | | | | | |
| PMC-LLaMA | 7B/ 13B | MedQA<br>MedMCQA | 20K<br>183K | 4.7M<br>26.7M | Llama 2 | LLaMA |
| Me-LLaMA | 13B | | | | | |
| Meditron | 7B/ 70B | | | | | |
| Med-Llama-3 | 8B | | | | Llama 3 | |
| Llama 3 Fine-tuned | 8B | | | | | |
| gpt-oss Fine-tuned | 20B | | | | gpt-oss | / |
| **Fine-tuned over clinical notes** | | | | | | |
| Llama 3 | 8B | Inpatient records from Yale New Haven Health System | 13K | 10.72M | Llama 3 | / |
| Med-Llama-3 | 8B | | | | Med-Llama-3 | |